\documentclass[runningheads]{llncs}

\usepackage[mobile]{eccv}

\usepackage{eccvabbrv}

\usepackage{graphicx}
\usepackage{booktabs}
\usepackage{wrapfig}
\usepackage[accsupp]{axessibility}  
\usepackage{algorithm}
\usepackage{algorithmic}
\usepackage{afterpage}
\usepackage{amsmath}
\usepackage{bm}

\usepackage{hyperref}
\usepackage{xcolor}
\usepackage{colortbl}
\usepackage{orcidlink}
\usepackage{microtype}
\usepackage{xcolor}
\definecolor{tabfirst}{rgb}{1, 0.7, 0.7}
\definecolor{tabsecond}{rgb}{1, 0.85, 0.7}
\definecolor{tabthird}{rgb}{1, 1, 0.7}

\usepackage{tabularx}
\usepackage{booktabs}

\begin{document}

\newcommand{\zjy}[1]{\textcolor{blue}{#1}}
\newcommand{\zhj}[1]{\textcolor{red}{#1}}
\newcommand{\zxy}[1]{\textcolor{cyan}{#1}}

\title{WildSplat: Feedforward Gaussian Splatting from Unposed In-the-Wild Images} 

\titlerunning{WildSplat}

\author{Xiyu Zhang\inst{1,2,*,\ddagger} \and
Jingyu Zhuang\inst{2,*} \and
Hongjia Zhai\inst{1} \and
Zizheng Yan\inst{2} \and \ \ \ \ \
Jinwei Chen\inst{2} \and
Guofeng Zhang\inst{1,\dagger} \and
Qingnan Fan\inst{2,\dagger}}

\authorrunning{X.~Zhang et al.}

\institute{State Key Lab of CAD\&CG, Zhejiang University, China \and
vivo BlueImage Lab, China\\
\textsuperscript{*}Equal contribution. \textsuperscript{\ensuremath{\dagger}}Corresponding author. \\ \textsuperscript{\ensuremath{\ddagger}}Work done during an internship at vivo BlueImage Lab.}

\maketitle
\noindent\begin{minipage}{\textwidth}
    \centering
    \includegraphics[width=1.0\linewidth]{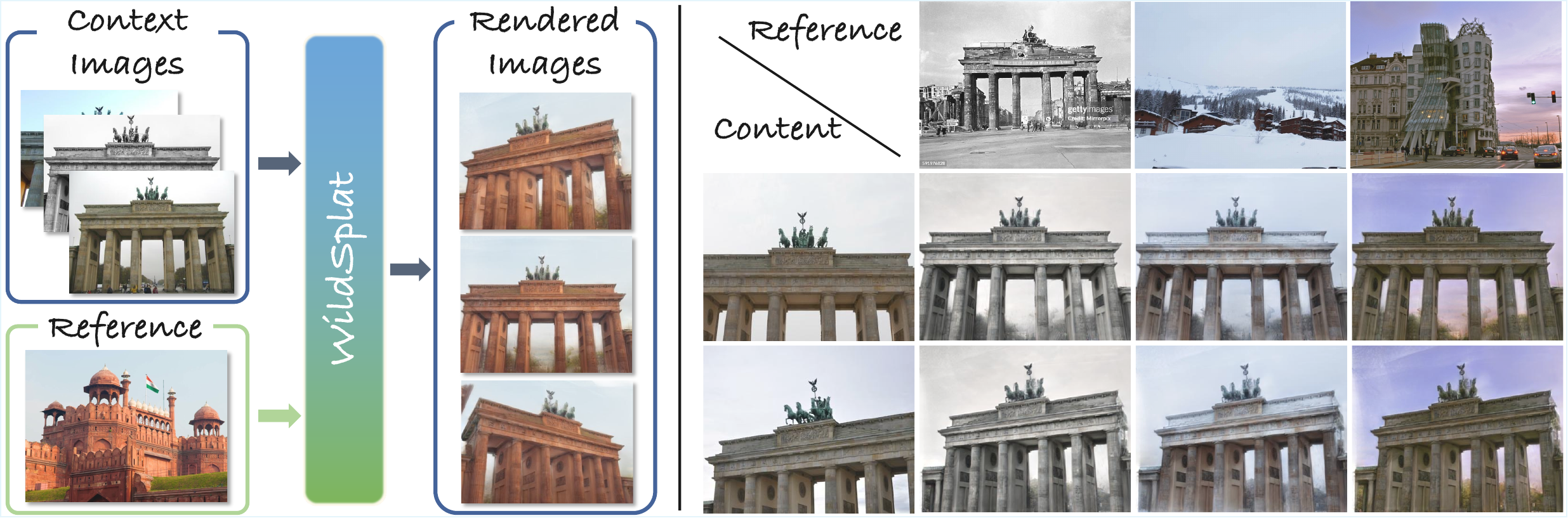}
    \captionsetup{type=figure,belowskip=0pt}
    \captionof{figure}{
    Given a set of unposed in-the-wild images with varying appearances and a reference image, \textbf{WildSplat} can reconstruct a high-quality 3D scene in a single feedforward pass. The reconstructed scene can then be rendered with the target appearance specified by the reference image. 
    }
    \label{fig:teaser}
\end{minipage}
\begin{abstract}
While feedforward 3D reconstruction excels at efficient novel view synthesis, it typically falters when faced with scenes under varying illumination. To this end, we
introduce \textbf{WildSplat}, the first feedforward 3D Gaussian Splatting framework capable of appearance-conditioned novel-view synthesis for unposed in-the-wild images. 
To handle inconsistent photometric conditions, we propose a dual-branch architecture that explicitly decouples geometry from appearance. The geometry branch extracts an appearance-invariant 3D structure and jointly predicts camera poses. To govern the rendering appearance, the appearance branch injects target appearance cues into the content features via a globally pre-modulated cross-attention mechanism. To further prevent feature entanglement, we introduce a joint multi-reference training strategy that stabilizes the training process. Extensive experiments show that WildSplat surpasses existing optimization-based and feedforward methods, achieving state-of-the-art performance in in-the-wild novel view synthesis and appearance editing from sparse inputs in a single forward pass.
\keywords{Novel View Synthesis \and 3D Gaussian Splatting \and Feedforward Reconstruction \and In-the-Wild}
\end{abstract}

\section{Introduction}
\label{sec:intro}
Novel view synthesis from multiple images is a longstanding problem in computer vision and graphics, forming the foundation for many downstream applications, such as virtual reality, augmented reality, and robotics.
Recently, Neural Radiance Fields (NeRF)~\cite{mildenhall2021nerf} and 3D Gaussian Splatting (3DGS)~\cite{kerbl20233d} have achieved photorealistic rendering by representing scenes with neural networks or 3D Gaussians.
However, these methods assume static scene geometry, materials, and lighting—an assumption usually violated for in-the-wild photo collections.
Such images, typically sourced from the internet, capture the same scene under varying conditions, exhibiting significant appearance differences in lighting, weather, and camera exposure.

To handle such complex scenarios, the community has historically relied on optimization-based approaches. 
Pioneering methods like NeRF-W~\cite{martin2021nerf} and subsequent appearance-aware 3DGS techniques~\cite{zhang2024gaussian, kulhanek2024wildgaussians, xu2024wild, dahmani2024swag} address this by optimizing per-image appearance embeddings to decouple the shared scene structure from these photometric variations. 
While these methods achieve impressive appearance interpolation, they strictly require accurately precomputed camera poses, and their performance heavily depends on dense image collections and lengthy per-scene optimization.
When only sparse views are available, optimizing per-image embeddings becomes severely under-constrained, leading to inaccurate geometry and a complete lack of reconstruction efficiency.

Meanwhile, some methods leverage large-scale multi-view datasets to learn generalizable priors and predict 3D representations in a single forward pass~\cite{wang2024dust3r,wang2025vggt,jiang2025anysplat, ye2024noposplat}.
Without requiring exhaustive per-scene optimization, these feedforward methods enable efficient and photorealistic novel-view synthesis.
However, extending such models to in-the-wild photo collections remains highly challenging.
%
%
When the appearance of the input images varies significantly, existing feedforward frameworks struggle to disentangle underlying 3D geometry from illumination changes, resulting in noticeable appearance inconsistencies in the rendered results.

To this end, we introduce \textbf{WildSplat}, the first feedforward 3D Gaussian Splatting framework designed 
for in-the-wild scenes with varying appearances. 
By conditioning the reconstruction on a designated reference image, WildSplat enables photorealistic novel-view synthesis without exhaustive per-scene optimization.
Specifically, we propose a dual-branch architecture to explicitly decouple geometry from appearance. This design allows us to reconstruct 3D Gaussians by separating appearance-invariant geometry from target-conditioned appearance. The geometry branch is responsible for predicting this invariant geometry, alongside robust content features that encode the structural information of the scene.
To control the rendering appearance, an appearance encoder extracts features from the reference image and injects the appearance information into the content features via a cross-attention-based appearance injector.
Furthermore, to stabilize the training process, we introduce a multi-reference training paradigm. 
Within a single training batch, our framework constructs an appearance-invariant geometry but simultaneously renders it into multiple views under distinct appearance conditions. 
This strategy enables the model to effectively accommodate photometric variations, ensuring accurate appearance modeling and photorealistic rendering for diverse target views.

In summary, our main contributions are as follows:
\begin{enumerate}
    \item 
    We propose \textbf{WildSplat}, the first feedforward 3D Gaussian Splatting framework enabling appearance-conditioned novel-view synthesis from sparse, unposed, in-the-wild image collections.
    \item We design a novel architecture that explicitly injects target appearance features into the content features from the geometry backbone via global pre-modulated cross-attention, enabling appearance-conditioned synthesis in a single forward pass.
    \item We introduce a multi-reference training paradigm tailored for photometrically inconsistent data, encouraging the model to learn robust 3D priors and accurately model real-world appearances.
    \item 
    We conduct extensive experiments to demonstrate that WildSplat achieves state-of-the-art performance on in-the-wild datasets, significantly outperforming both existing feedforward approaches and optimization-based methods.
\end{enumerate}

\section{Related Work}
\label{sec:related}
\subsection{Novel View Synthesis}

Novel view synthesis aims to infer scene structure and appearance based on a given set of input views.
Traditional works focus on explicit~\cite{waechter2014let, 1996photographs, liu2019soft} or discrete volumetric~\cite{kutulakos2000theory, seitz1999photorealistic, szeliski1999stereo} representations of the underlying geometry, while recent methods employ radiance field-based representations~\cite{mildenhall2021nerf} or 3D Gaussian Splatting~\cite{kerbl20233d} to achieve unprecedented photorealistic rendering quality.
Neural Radiance Fields (NeRF)~\cite{mildenhall2021nerf} and related methods~\cite{barron2021mip, barron2022mip, barron2023zip} have pioneered high-quality novel view synthesis by training neural networks to learn neural fields that implicitly represent scene geometry and appearance. However, their reliance on computationally expensive volume rendering limits real-time performance.
%
In contrast, 3D Gaussian Splatting (3DGS)~\cite{kerbl20233d} models a scene as a large collection of learnable anisotropic Gaussian primitives.
Its explicit and efficient representation has further benefited downstream tasks such as visual localization, surface reconstruction, and open-vocabulary panoptic mapping~\cite{zhai2025splatloc, zhang2026atlasgs, zhai2026onlinepg}.

The aforementioned novel view synthesis methods assume that scene geometry, materials, and lighting are static; however, in-the-wild images collected from the Internet often violate this assumption.
To address this issue, NeRF-W~\cite{martin2021nerf} learns per-image appearance and transient embeddings to model photometric variations and disentangle transient elements.
GS-W~\cite{zhang2024gaussian} employs 3DGS to represent the scene and introduces separate intrinsic and dynamic features for each Gaussian to capture the static scene along with dynamic variations.
WildGaussians~\cite{kulhanek2024wildgaussians} leverages robust DINO~\cite{oquab2023dinov2} features to eliminate occluders and integrates an appearance modeling module within 3DGS to predict color affine transformations.
Similarly, Splatfacto-W~\cite{xu2024splatfacto} integrates per-image appearance embeddings and per-Gaussian color features into the rasterization process.
Furthermore, Wild-GS~\cite{xu2024wild} decomposes appearance into image-based global and local embeddings and leverages a triplane representation to achieve explicit local appearance modeling.
Despite these advances, most in-the-wild NeRF and 3DGS methods typically rely on Structure-from-Motion tools~\cite{schonberger2016structure} to obtain camera poses and still involve redundant optimization processes, which significantly reduce reconstruction efficiency.

\subsection{Feedforward Reconstruction}

Since gradient backpropagation in per-scene optimization is computationally expensive and time-consuming, some methods attempt to learn powerful priors from large-scale datasets to reconstruct scenes and enable novel view synthesis via a single feedforward pass.
pixelNeRF~\cite{yu2021pixelnerf} introduces a fully convolutional architecture that conditions on input images to predict pixel-aligned features for radiance field reconstruction.
Subsequent feedforward NeRF models further improve performance by incorporating feature matching~\cite{chen2021mvsnerf, chen2025explicit}, Transformers~\cite{du2023learning, miyato2023gta, sajjadi2022scene}, and 3D volumetric representations~\cite{xu2024murf}. 
However, due to the inherently pixel-wise volume rendering, existing feedforward NeRF models still suffer from slow rendering speeds.

Recently, a growing number of feedforward models based on 3DGS have been proposed to avoid the expensive volume rendering in NeRF.
pixelSplat~\cite{charatan2024pixelsplat} learns a Transformer-based model to regress Gaussian parameters from two input views by extracting cross-view-aware features and predicting probabilistic depth distributions for depth sampling.
MVSplat~\cite{chen2024mvsplat} instead introduces a cost-volume formulation that aggregates cross-view matching cues, leading to more accurate geometry estimation.
%
Without requiring accurate camera poses as input, NoPoSplat~\cite{ye2024noposplat} anchors one input camera coordinate system as the canonical space, predicts Gaussians for all views within this space, and encodes camera intrinsics as token embeddings to enable accurate scene scale prediction.
AnySplat~\cite{jiang2025anysplat} distills camera and geometry priors from a pretrained VGGT~\cite{wang2025vggt} model to encode input images into high-dimensional features, which are then decoded into Gaussian parameters and camera poses.
Despite the impressive progress of feedforward reconstruction models, they rely on a strict assumption of photometric consistency across input views and struggle to disentangle the underlying 3D geometry from illumination variations. 
When in-the-wild images are used as inputs, appearance inconsistencies caused by varying weather and lighting conditions can severely degrade the visual consistency of the reconstructed scenes produced by existing feedforward methods.

\section{Methods}
\label{sec:methods}
\begin{figure}[t]
    \centering
    \includegraphics[width=1.0\linewidth]{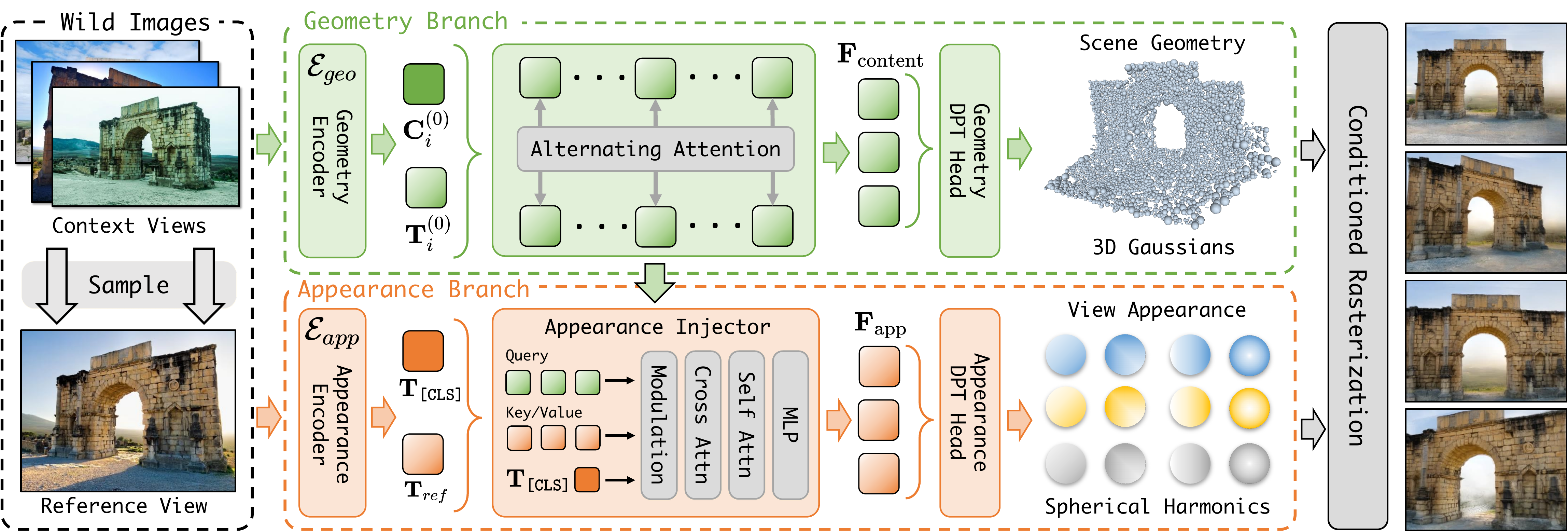}
    \caption{\textbf{Overview of WildSplat}. Given context images with varying appearances and a reference view, the Geometry Branch extracts content features and predicts scene geometry along with explicit 3D Gaussian attributes. Concurrently, the Appearance Branch employs an Appearance Injector, conditioned on a reference appearance image, to predict view-consistent color attributes. The decoupled geometry and conditioned appearance are then combined for conditioned rasterization to synthesize novel views with the designated reference appearance.}
    \label{fig:pipeline}
\end{figure}

In~\cref{fig:pipeline}, we present \textbf{WildSplat}, a novel feedforward 3D Gaussian Splatting framework designed for unposed scenes with varying appearances. 
Given a set of context images exhibiting various appearances and a designated reference image, WildSplat decouples the scene into appearance-invariant geometry and conditioned appearance under the target appearance.
The problem formulation is introduced in~\cref{sec:setup}, followed by a detailed description of our WildSplat in~\cref{sec:model}. Finally, the multi-reference training strategy is presented in~\cref{sec:training}.

\subsection{Problem formulation}
\label{sec:setup}
Unlike standard novel view synthesis settings that rely on strict multi-view photometric consistency, we focus on unposed in-the-wild scenarios where the context images $\{I_i\}$ frequently exhibit photometric variations. 
To resolve the resulting photometric ambiguities, our framework explicitly decouples the geometry and appearance and conditions the appearance on a target reference image $I_{ref}$, which provides the designated appearance cues. 
Consequently, WildSplat jointly predicts the camera poses $\{p_i\}$ of the input images, the underlying 3D Gaussian geometry attributes $\mathcal G$, and the conditioned appearance attributes $\mathcal{C}_{ref}$. 
This formulation ensures that the reconstructed 3D representation geometrically aligns with the context views while strictly matching the appearance of $I_{ref}$. 
Formally, WildSplat is defined as:
\begin{align}
f: \{I_i\}, I_{ref} \mapsto (\mathcal G, \mathcal{C}_{ref}, \{p_i\}),
\end{align}
where $\mathcal G$ consists of Gaussian positions $\bm\mu \in \mathbb{R}^{G\times 3}$, rotations $\mathbf q\in \mathbb{R}^{G\times 4}$, scales $\mathbf s\in \mathbb{R}^{G\times 3}$, and opacities $\bm\alpha \in \mathbb{R}^G$. $\mathcal{C}_{ref} \in \mathbb{R}^{G\times 3\times(k+1)^2}$ contains spherical harmonics coefficients of degree $k$ conditioned on the reference image $I_{ref}$, and $G$ is the number of Gaussians.

\subsection{Decoupled Geometry and Appearance Modeling}
\label{sec:model}
Previous feedforward methods typically predict 3D Gaussian geometry and color attributes jointly from an entangled representation. To prevent the target appearance conditioning from inadvertently altering the underlying geometric structure, we explicitly decouple these properties and design a dual-branch framework to separate the prediction of these attributes. 
Specifically, the geometry branch extracts appearance-invariant geometry that remains consistent across arbitrary reference images, while the appearance branch, via an appearance injector module, predicts decoupled appearance from the conditioned reference image.

\noindent\textbf{Geometry Branch.} We build our geometry branch upon the VGGT-like architecture~\cite{wang2025vggt}.
Initially, $N$ context images are processed independently by a DINOv2~\cite{oquab2023dinov2} based geometry encoder $\mathcal{E}_{\text{geo}}$, yielding dense patch-level tokens $\mathbf{T}_i^{(0)} = \mathcal{E}_{\text{geo}}(I_i) \in \mathbb{R}^{P \times D}$ for each view $i \in \{1, \dots, N\}$. 
To enable camera pose estimation, a set of learnable camera tokens $\mathbf{C}^{(0)}_i \in \mathbb{R}^{5\times D}$ is concatenated with the patch tokens. These combined tokens are refined through $L$ alternating attention blocks. Let $\mathbf{T}_i^{(k)}$ and $\mathbf{C}_i^{(k)}$ denote the patch and camera tokens for the $i$-th view after the $k$-th layer, respectively.

To encapsulate the appearance-invariant structural information of the scene, we extract and aggregate multi-scale representations from a subset of transformer layers $\mathcal{K} = \{l_1, \dots, l_K\}$. We formally define these aggregated representations as the content features, which serve as a foundational geometric scaffold prior to any appearance conditioning:
\begin{equation}
\mathbf{F}_{\text{content}} = \left\{ \mathbf{T}_i^{(k)} \mid k \in \mathcal{K}, i=1 \dots N  \right\}.
\end{equation}

Based on these enriched representations, we employ specialized heads to regress 3D Gaussian properties. Specifically, DPT-based heads $\mathcal{H}$ map the aggregated features $\mathbf{F}_{\text{content}}$ to pixel-aligned depth maps $\{d_i\}$ and Gaussian attributes $\{s_i, q_i, \alpha_i\}$. Simultaneously, the refined camera tokens $\mathbf{C}_i^{(L)}$ are processed by a camera head to predict poses $\{p_i\}$. The explicit 3D Gaussian centers $\mu_i$ are then derived via a back-projection layer:
\begin{equation}
\begin{aligned}
\{d_i\} = \mathcal{H}_{\text{depth}}(\mathbf{F}_{\text{content}}), & \quad p_i = \mathcal{H}_{\text{cam}}(\mathbf{C}_i^{(L)}), \\
\{s_i, q_i, \alpha_i\} = \mathcal{H}_{\text{attr}}(\mathbf{F}_{\text{content}}), & \quad \mu_i = \text{backproj}(d_i, p_i).
\end{aligned}
\end{equation}
Finally, based on the predicted camera pose, we transform the 3D Gaussians from the viewpoint into the reconstructed global coordinates, thereby obtaining a viewpoint-independent representation of the scene geometry: $\mathcal{G}=\{\bm{\mu}, \mathbf{s}, \mathbf{q}, \bm{\alpha}\}$.

\noindent\textbf{Appearance Branch.}
The appearance branch aims to predict appearance conditioned on the target reference image.
Given the target reference image $I_{ref}$, we first employ another DINOv2 based appearance encoder $\mathcal{E}_{app}$ to extract the appearance information:
\begin{equation}
    \mathbf{T}_{\texttt{[CLS]}}, \mathbf{T}_{ref} = \mathcal{E}_{app}(I_{ref}),
\end{equation}
where $\mathbf T_{\texttt{[CLS]}} \in \mathbb{R}^{D}$ corresponds to the \texttt{[CLS]} token that represents global appearance, and $\mathbf T_{ref} \in \mathbb{R}^{P_a \times {D}}$ represents local dense appearance tokens, with $P_a$ being the number of patches and ${D}$ the feature dimension.
Then, we use a cross-attention mechanism to integrate the target appearance into the content features $\mathbf{F}_\text{content}$. 
The content features $\mathbf{F}_\text{content}$ extracted from the geometry branch act as the queries ($Q$), while the reference appearance tokens $\mathbf T_{ref}$ serve as the keys ($K$) and values ($V$). 

However, the content features $\mathbf{F}_\text{content}$ from the geometry branch inherently retain the original photometric properties of the context views. 
To explicitly mitigate the influence of the original appearances, we introduce a global pre-modulation step before the cross-attention operation.
We leverage the global descriptor $\mathbf T_{\texttt{[CLS]}}$ to holistically shift the content features into the target appearance domain. 
Specifically, an AdaLN-Zero layer~\cite{peebles2023scalable} is employed to decode $\mathbf T_{\texttt{[CLS]}}$ into three parameters: a scale $\gamma$, a shift $\beta$, and an adaptive gating $\eta$. These parameters are then used to modulate the appearance information within the content feature $\mathbf F_{\text{content}}$ as follows:
\begin{equation}
    \tilde{\mathbf{T}}_i^{(k)} = \gamma \odot \text{LayerNorm}(\mathbf{T}_i^{(k)}) + \beta, \quad [\gamma, \beta, \eta] = \text{MLP}_{\text{cond}}(\mathbf{T}_{\texttt{[CLS]}}),
\end{equation}
where $\odot$ denotes element-wise multiplication.
The appearance-modulated feature $\tilde{T}^{(k)}$ is then utilized as the query in the cross-attention layer to aggregate fine-grained, condition-specific details from $T_{ref}$. 
This is followed by a self-attention layer to ensure spatial regularization. The complete decoupled injection process is formulated as:
\begin{eqnarray}
& \bar{\mathbf{T}}_i^{(k)} = \mathbf{T}_i^{(k)} + \eta \odot \text{CrossAttn}(\tilde{\mathbf{T}}_i^{(k)}, \mathbf{T}_{ref}), \hat{\mathbf{T}}_i^{(k)} = \bar{\mathbf{T}}_i^{(k)} + \text{SelfAttn}(\bar{\mathbf{T}}_i^{(k)}),&\\
&\mathbf{T}_{\text{app}, i}^{(k)} = \hat{\mathbf{T}}_i^{(k)} + \text{MLP}(\hat{\mathbf{T}}_i^{(k)}).&
\end{eqnarray}
Finally, the updated appearance features $\mathbf{F}_{\text{app}} = \{ \mathbf{T}_{\text{app}, i}^{(k)} \mid k \in \mathcal{K}, i=1, \dots, N \}$ are routed to the appearance head $\mathcal{H}_{\text{color}}$ to predict the conditioned 3D Gaussian color attributes $\mathcal{C}_{ref}$:
\begin{equation}
\mathcal{C}_{ref} = \mathcal{H}_{\text{color}}(\mathbf{F}_{\text{app}}).
\end{equation}
Combined with the appearance-invariant geometry attributes $\mathcal{G}=\{\bm{\mu}, \mathbf{s}, \mathbf{q}, \bm{\alpha}\}$, the fully decoupled 3D representation $\{\mathcal G, \mathcal{C}_{ref}\}$ is ready for conditioned rasterization.

\noindent\textbf{Conditioned Rasterization.}
With the decoupled geometry and appearance attributes, we perform a conditioned rasterization to ensure the rendered novel views maintain a globally consistent 3D structure while seamlessly integrating the conditioned appearance.
Formally, given the decoupled 3D representation $\{\mathcal G, \mathcal{C}_{ref}\}$, and camera poses $\{p_i\}$ of target views, the conditioned representation is rendered to synthesize the predicted outputs $\{\hat{I}_i\}$:
\begin{equation}
\{\hat{I}_i\} = \mathcal{R}({\mathcal G, \mathcal{C}_{ref}}, \{p_i\}),
\end{equation}
where $\mathcal{R}$ denotes the tile-based differentiable rasterization process of 3D Gaussian Splatting.

\subsection{Training Paradigm}
\label{sec:training}
\noindent\textbf{Geometry-guided View Sampling.}
In feedforward training, input views must share sufficient visual overlap to establish reliable geometric correspondences. 
Standard sampling strategies, however, rely solely on camera extrinsics or sequential information. 
These approaches may fail on in-the-wild datasets, where varying focal lengths mean that even minor pose shifts can drastically alter the captured content. 
To guarantee a valid multi-view context, we introduce an overlap-based sampling strategy. 
Leveraging SfM points, we formulate an overlap matrix and a scale matrix to quantify the ratio of shared 3D points and the minimum depth ratio between views, respectively. 
Based on predefined thresholds, we first discard view pairs lacking adequate spatial overlap or scale consistency. 
Then, initializing from a strongly connected candidate view, we iteratively aggregate valid neighboring views. The complete procedure is detailed in~\cref{algo:sampling}.
\begin{algorithm}[t]
\caption{Geometry-guided View Sampling}
\label{algo:sampling}
\textbf{Input:} Matrices $S$ (scale) \& $R$ (ratio), num views $N$, sample count $K$, thresholds $\tau_{s}, \tau_{r}, \tau_{n}$ \\
\textbf{Output:} Selected indices $I$
\begin{algorithmic}[1]
\STATE $V \leftarrow (S \geq \tau_{s}) \land (R \geq \tau_{r})$ \hfill \textit{// Valid mask}
\STATE $C \leftarrow \{ i \mid \sum_j V_{i,j} > \tau_{n} \}$ \hfill \textit{// Valid starting rows}
\STATE Initialize queue $Q \leftarrow [c]$ and set $I \leftarrow \{c\}$ where $c \sim \text{Uniform}(C)$
\WHILE{$|I| < N$ \AND $Q \neq \emptyset$}
    \STATE $u \leftarrow Q.\text{pop}()$
    \STATE $S_k \leftarrow \text{Sample up to } K \text{ elements from } \{ v \mid V_{u,v} \land v \notin I\}$
    \STATE $Q.\text{push}(S_k)$ \AND $I \leftarrow I \cup S_k$
\ENDWHILE
\RETURN $I$
\end{algorithmic}
\end{algorithm} 

\noindent\textbf{Joint Multi-Reference Supervision.}
To stabilize the training process and explicitly prevent the entanglement of geometry and appearance, we introduce a joint multi-reference training strategy. 
During a training iteration, after extracting the appearance-invariant geometry $\mathcal G$, the intermediate content features $\mathbf F_\text{content}$, and the camera poses $\{p_i\}$ from a batch of sampled multi-view images, we explicitly supervise the appearance conditioning. 
To capture the diverse appearances within the batch, we randomly sample a subset of $M$ images, $\{I_m\}$, as target appearance references and ground truth.
The appearance branch processes these $M$ references to extract conditioned appearance attributes $\{\mathcal{C}_{ref,m}\}$. 
Finally, we render images from the viewpoint of reference images 
explicitly conditioned on their respective appearance information.
The rendered $M$ images $\{\hat{I}_m\}$ are subsequently used for end-to-end supervision.
To prevent appearance leakage and ensure the geometry remains illumination-agnostic, we apply color jittering to the input images and random cropping to the references prior to feature extraction.

\noindent\textbf{Training Objectives.}
The joint objective function is defined as the summation of the reconstruction losses over these rendered images:
\begin{equation}
\mathcal{L}_\text{total} = \mathcal{L}_\text{rgb} + \lambda_1\mathcal{L}_\text{pose}.
\end{equation}
$\mathcal{L}_\text{rgb}$ is the photometric loss, which is formulated as a weighted combination of multiple loss terms:
\begin{equation}
\mathcal{L}_\text{rgb} = \lambda_{2}\mathcal{L}_\text{MSE} + \lambda_{3} \mathcal{L}_\text{SSIM} + \lambda_{4} \mathcal{L}_\text{LPIPS},
\end{equation}
$\mathcal{L}_\text{pose}$ is the distillation loss in~\cite{jiang2025anysplat} to regularize the camera pose estimation and mitigate the adverse effects of inconsistent inputs during training. The corresponding balancing weights are set to $\lambda_i = \{10, 1.0, 0.05, 0.05\}$, respectively.

\begin{table}[t]
\centering
\small
\setlength{\tabcolsep}{4pt} 
\renewcommand{\arraystretch}{1.1} 
\caption{\textbf{Quantitative comparison} of novel view synthesis on MegaScenes~\cite{tung2024megascenes}.  We evaluate models under sparse input configurations (4, 8, and 12 views) using a fixed test set. Performance is measured using PSNR ($\uparrow$), SSIM ($\uparrow$), and LPIPS ($\downarrow$).}
\label{tab:comp_nvs_megascenes}
\resizebox{0.9\linewidth}{!}{
\begin{tabular}{@{}lccccccccc@{}}
\toprule
Method& \multicolumn{3}{c}{4 views} & \multicolumn{3}{c}{8 views} & \multicolumn{3}{c}{12 views} \\
\midrule
 & PSNR$\uparrow$ & SSIM$\uparrow$ & LPIPS$\downarrow$ & PSNR$\uparrow$ & SSIM$\uparrow$ & LPIPS$\downarrow$ & PSNR$\uparrow$ & SSIM$\uparrow$ & LPIPS$\downarrow$ \\
\midrule
FSGS~\cite{zhu2024fsgs}&12.88&0.352&0.463&13.28&0.371&0.454&13.43&0.376&0.453 \\
GS-W~\cite{zhang2024gaussian}&9.54&0.265&0.539&10.76&0.294&0.500&9.81&0.311&0.526 \\
WildGaussian~\cite{kulhanek2024wildgaussians}&14.38&0.385&0.458&15.32&0.436&0.433&15.68&0.448&0.452 \\
AnySplat~\cite{jiang2025anysplat}&\cellcolor{tabthird}14.58&\cellcolor{tabthird}0.559&\cellcolor{tabthird}0.313&\cellcolor{tabsecond}16.16&\cellcolor{tabthird}0.594&\cellcolor{tabthird}0.274&\cellcolor{tabsecond}16.31&\cellcolor{tabthird}0.601&\cellcolor{tabthird}0.273 \\
WorldMirror~\cite{liu2025worldmirror} & \cellcolor{tabsecond}15.33&\cellcolor{tabsecond}0.609&\cellcolor{tabsecond}0.288 & \cellcolor{tabthird}16.08&\cellcolor{tabsecond}0.633&\cellcolor{tabsecond}0.281& \cellcolor{tabthird}16.05&\cellcolor{tabsecond}0.629&\cellcolor{tabsecond}0.287 \\
Ours         & \cellcolor{tabfirst}17.91 & \cellcolor{tabfirst}0.635 & \cellcolor{tabfirst}0.235 & \cellcolor{tabfirst}18.65 & \cellcolor{tabfirst}0.668 & \cellcolor{tabfirst}0.218 & \cellcolor{tabfirst}19.20 & \cellcolor{tabfirst}0.683 & \cellcolor{tabfirst}0.210 \\
\bottomrule
\end{tabular}
}
\end{table}

\begin{table}[t]
\centering
\small
\setlength{\tabcolsep}{4pt} 
\renewcommand{\arraystretch}{1.15} 
\caption{\textbf{Quantitative comparison} of novel view synthesis on Phototourism~\cite{snavely2006photo}.  We evaluate models under sparse input configurations (4, 8, and 12 views) using a fixed test set. Performance is measured using PSNR ($\uparrow$), SSIM ($\uparrow$), and LPIPS ($\downarrow$).}
\label{tab:comp_nvs_phototourism}
\resizebox{0.9\linewidth}{!}{
\begin{tabular}{@{}lccccccccc@{}}
\toprule
Method & \multicolumn{3}{c}{4 views} & \multicolumn{3}{c}{8 views} & \multicolumn{3}{c}{12 views} \\
\midrule
 & PSNR$\uparrow$ & SSIM$\uparrow$ & LPIPS$\downarrow$ & PSNR$\uparrow$ & SSIM$\uparrow$ & LPIPS$\downarrow$ & PSNR$\uparrow$ & SSIM$\uparrow$ & LPIPS$\downarrow$ \\
\midrule
FSGS~\cite{zhu2024fsgs}&13.63&0.476&0.413&14.85&0.532&0.366&15.39&0.573&0.346 \\
GS-W~\cite{zhang2024gaussian}&9.40&0.344&0.512&10.19&0.354&0.490&12.63&0.399&0.437 \\
WildGaussian~\cite{kulhanek2024wildgaussians}&14.95&0.485&0.436&\cellcolor{tabthird}17.02&0.567&0.371&\cellcolor{tabthird}17.65&0.616&0.353 \\
AnySplat~\cite{jiang2025anysplat}&\cellcolor{tabsecond}15.61&\cellcolor{tabthird}0.616&\cellcolor{tabthird}0.279&\cellcolor{tabsecond}17.35&\cellcolor{tabthird}0.681&\cellcolor{tabthird}0.240&\cellcolor{tabsecond}18.00&\cellcolor{tabthird}0.700&\cellcolor{tabsecond}0.228 \\
WorldMirror~\cite{liu2025worldmirror}&\cellcolor{tabthird}15.38&\cellcolor{tabsecond}0.664&\cellcolor{tabsecond}0.273&16.71&\cellcolor{tabsecond}0.699&\cellcolor{tabthird}0.252& 17.24&\cellcolor{tabsecond}0.724&\cellcolor{tabthird}0.234 \\

Ours & \cellcolor{tabfirst}19.57& \cellcolor{tabfirst} 0.727&  \cellcolor{tabfirst} 0.178 & \cellcolor{tabfirst}21.15&  \cellcolor{tabfirst} 0.771&  \cellcolor{tabfirst} 0.146 & 
\cellcolor{tabfirst}21.47&  \cellcolor{tabfirst} 0.788& \cellcolor{tabfirst} 0.137 \\
\bottomrule
\end{tabular}
}
\end{table}

\section{Experiments}
\label{sec:exps}
\subsection{Experimental Setup}

\begin{figure}[t]
    \centering
    \includegraphics[width=1.0\linewidth]{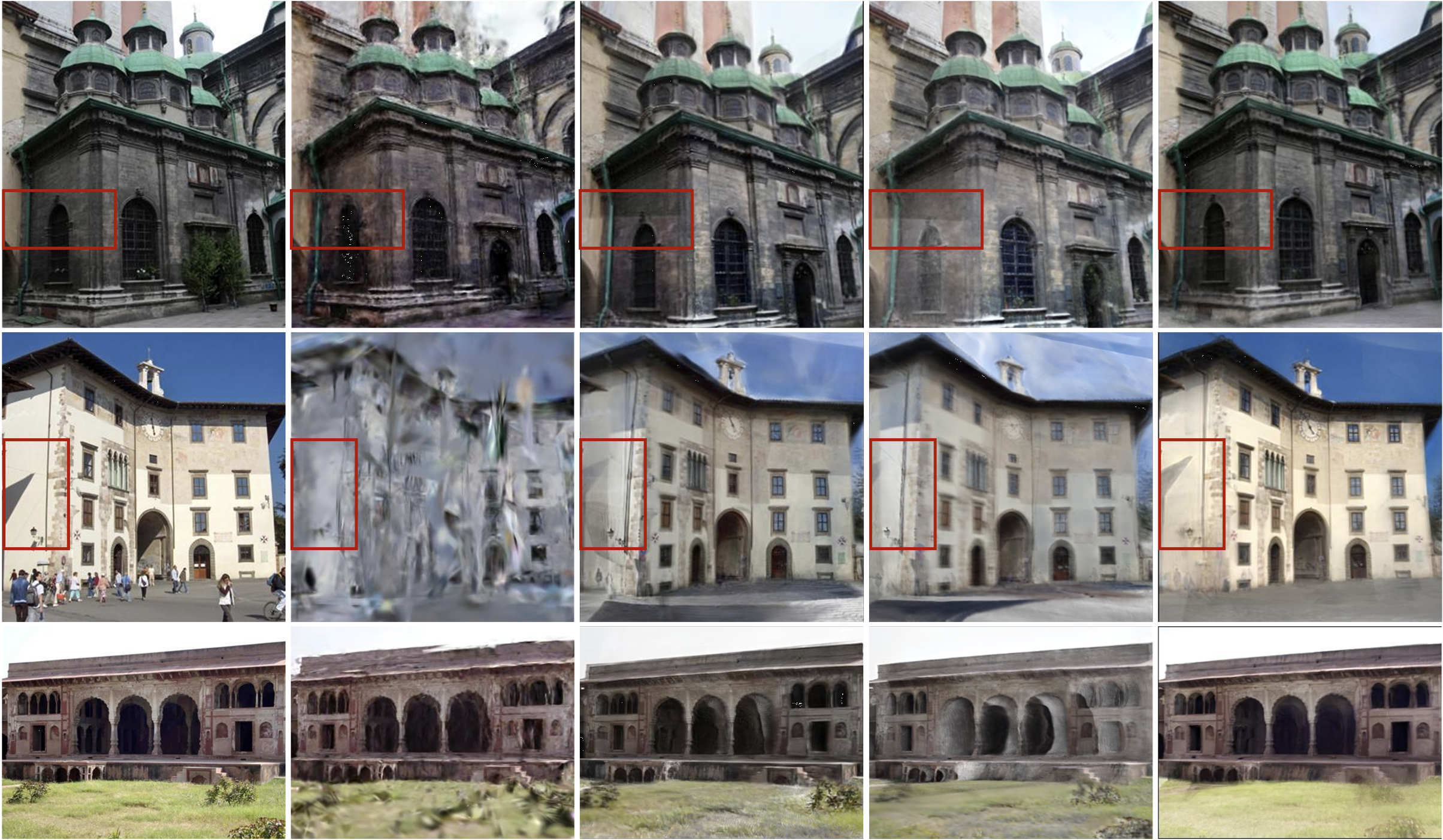}
    \begin{tabular}{@{}c@{\hspace{0.02\linewidth}}c@{\hspace{0.02\linewidth}}c@{\hspace{0.02\linewidth}}c@{\hspace{0.02\linewidth}}c@{}}
        \makebox[0.18\linewidth]{\scriptsize GT} & 
        \makebox[0.18\linewidth]{\scriptsize WildGaussian~\cite{kulhanek2024wildgaussians}} & 
        \makebox[0.18\linewidth]{\scriptsize AnySplat~\cite{jiang2025anysplat}} & 
        \makebox[0.18\linewidth]{\scriptsize WorldMirror~\cite{liu2025worldmirror}} & 
        \makebox[0.18\linewidth]{\scriptsize \textbf{Ours}} 
    \end{tabular}
    \caption{\textbf{Qualitative comparison} on the MegaScenes dataset~\cite{tung2024megascenes}.  We visualize novel view synthesis results under the 12-view setting. 
    Due to sparse input views, the optimization-based method WildGaussian produces many artifacts.
    Feedforward methods (AnySplat and WorldMirror) synthesize building exteriors with inconsistent colors (as highlighted in the red box) due to varying lighting conditions in the input images.
    In contrast, our method reconstructs realistic renderings with consistent appearance.
    }
    \label{fig:megascenes}
\end{figure}

\noindent\textbf{Dataset.}
We categorize our training datasets into two types: consistent appearance and diverse appearance. 
To establish geometric priors, we use the DL3DV dataset~\cite{ling2024dl3dv}, which provides strict photometric consistency. 
For diverse appearance, we incorporate MegaScenes~\cite{tung2024megascenes} and MegaDepth~\cite{li2018megadepth}.
Because these in-the-wild datasets lack multiple appearances from identical viewpoints, we augment sampled videos from DL3DV using an off-the-shelf video relighting model~\cite{liu2025tc}. 
This generates a multi-view, multi-appearance dataset with paired data under various synthetic lighting conditions, facilitating appearance disentanglement. 
For view sampling, we adopt interval-based sampling for DL3DV and geometry-guided view sampling for MegaDepth and MegaScenes.

To evaluate novel-view synthesis in scenes exhibiting complex appearance variations, following previous works~\cite{martin2021nerf, zhang2024gaussian, kulhanek2024wildgaussians}, we test our method on four scenes from the Phototourism~\cite{snavely2006photo} dataset.
In addition, we further use ten scenes sampled from the MegaScenes~\cite{tung2024megascenes} test set to increase scene diversity.
All evaluations are performed under 4-, 8-, and 12-view configurations using a fixed test set. 

\begin{figure}[t!]
    \centering
    \includegraphics[width=1.0\linewidth]{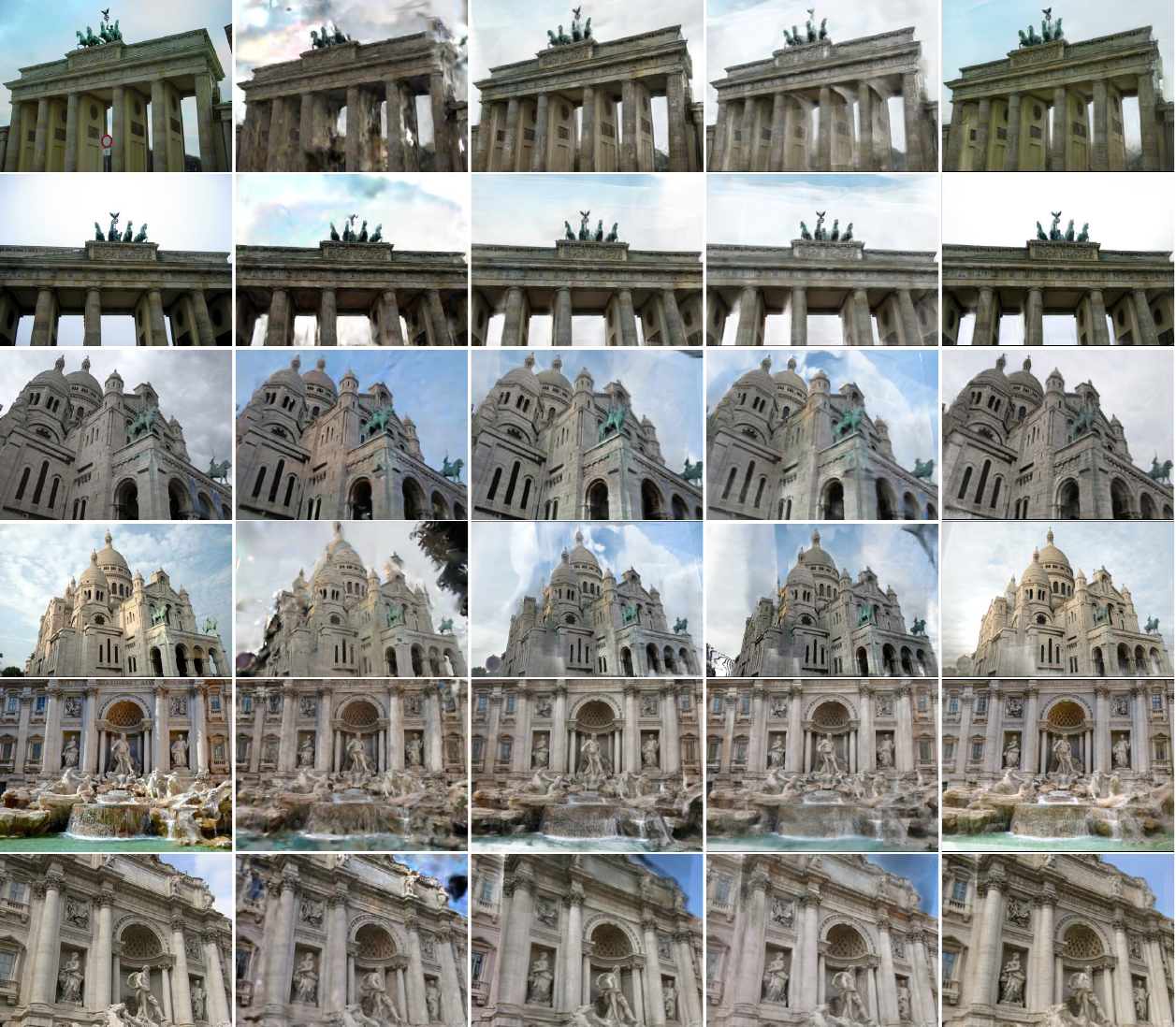}
    \begin{tabular}{@{}c@{\hspace{0.02\linewidth}}c@{\hspace{0.02\linewidth}}c@{\hspace{0.02\linewidth}}c@{\hspace{0.02\linewidth}}c@{}}
        \makebox[0.18\linewidth]{\scriptsize GT} & 
        \makebox[0.18\linewidth]{\scriptsize WildGaussian~\cite{kulhanek2024wildgaussians}} & 
        \makebox[0.18\linewidth]{\scriptsize AnySplat~\cite{jiang2025anysplat}} & 
        \makebox[0.18\linewidth]{\scriptsize WorldMirror~\cite{liu2025worldmirror}} & 
        \makebox[0.18\linewidth]{\scriptsize \textbf{Ours}} 
    \end{tabular}
    \caption{
    \textbf{Qualitative comparison} on the Phototourism dataset~\cite{snavely2006photo}.
    We visualize novel-view synthesis results under the 12-view setting.
    Compared with existing methods, our method successfully renders photorealistic and appearance-consistent images.
    }
    \label{fig:phototourism}
\end{figure}

\noindent\textbf{Baselines.}
We compare our approach against recent state-of-the-art methods based on Gaussian Splatting.
We categorize these baselines into two main paradigms: feedforward and optimization-based methods.
For feedforward approaches, we evaluate against AnySplat~\cite{jiang2025anysplat} and WorldMirror~\cite{liu2025worldmirror}. At evaluation time, we employ an evaluation-time pose alignment~\cite{ye2024noposplat} in which the Gaussian parameters are kept fixed while the target camera pose is optimized using a photometric loss, applied consistently to both the baselines and our method.
For optimization-based methods, we consider WildGaussian~\cite{kulhanek2024wildgaussians}, GS-W~\cite{zhang2024gaussian}, and FSGS~\cite{zhu2024fsgs}. Since these optimization-based approaches explicitly depend on accurate camera poses and dense point clouds, we use VGGT~\cite{wang2025vggt} to estimate the required geometric priors. We further ensure that all target views are correctly aligned to the input image coordinate system to enable a fair and controlled comparison. To prevent photometric information leakage, we follow the evaluation protocol established by NeRF-W~\cite{martin2021nerf}. Specifically, we use the left half of the ground-truth image as the reference and compute quantitative metrics solely on the right half.

\noindent\textbf{Implementation details.}
We implement our framework in PyTorch using the Adam optimizer. 
Our geometry branch comprises a DINOv2-large encoder ($D=1024$) and $L=24$ alternating-attention layers. The appearance branch consists of a pretrained DINOv2-base encoder coupled with a 4-layer appearance injector. For initialization, we load pretrained VGGT weights into the geometry backbone, apply zero-initialization to the AdaLN-Zero layers, and initialize all remaining parameters randomly. 
We train the model across four NVIDIA RTX H20 GPUs in two stages. First, we conduct a 30k-iteration warm-up on the DL3DV dataset with the appearance injector disabled to stabilize the Gaussian attribute heads. This is followed by 60k iterations of end-to-end training on the complete dataset. More experimental details are provided in the supplementary material.

\subsection{Main Results}
\noindent\textbf{Qualitative Results.}
Fig.~\ref{fig:megascenes} and Fig.~\ref{fig:phototourism} show the visual comparisons between our method and the baselines on MegaScenes~\cite{tung2024megascenes} and Phototourism~\cite{snavely2006photo}, respectively. 
The results demonstrate that, given a single reference image for appearance guidance, our method achieves the best rendering performance with consistent appearance.
Although WildGaussian can optimize an appearance embedding using a reference image, it produces many artifacts under sparse input views and requires long per-scene optimization times.
Both AnySplat and WorldMirror are feedforward methods that cannot decouple geometry and appearance. Consequently, due to varying lighting conditions in the input images, the reconstructed buildings exhibit mixed and inconsistent appearances.

\begin{table}[t!]
\centering
\small
\setlength{\tabcolsep}{2pt} 
\renewcommand{\arraystretch}{1.15} 
\caption{\textbf{Ablation Study.} We conduct an ablation study on the Phototourism dataset~\cite{snavely2006photo} to evaluate the contribution of individual components. Quantitative results, including PSNR ($\uparrow$), SSIM ($\uparrow$), and LPIPS ($\downarrow$), are reported for sparse input configurations of 4, 8, and 12 views.}
\label{tab:ablation}
\resizebox{0.9\linewidth}{!}{
\begin{tabular}{@{}lccccccccc@{}}
\toprule
{\footnotesize Setting} & \multicolumn{3}{c}{4 views} & \multicolumn{3}{c}{8 views} & \multicolumn{3}{c}{12 views} \\
\midrule
 & PSNR$\uparrow$ & SSIM$\uparrow$ & LPIPS$\downarrow$ & PSNR$\uparrow$ & SSIM$\uparrow$ & LPIPS$\downarrow$ & PSNR$\uparrow$ & SSIM$\uparrow$ & LPIPS$\downarrow$ \\
\midrule
w/o Multi-Ref Supervision& 16.88&0.668&\cellcolor{tabthird}0.221 &  18.77&\cellcolor{tabthird}0.723&\cellcolor{tabsecond}0.184 & 	19.35&\cellcolor{tabthird}0.746&\cellcolor{tabthird}0.177 \\
w/o Geo-guided Sampling & \cellcolor{tabthird}17.99&0.671&\cellcolor{tabsecond}0.220& 	\cellcolor{tabthird}19.24&0.709&0.192& 	\cellcolor{tabthird}19.58&0.731&0.182 \\
w/o Pre-Modulation & \cellcolor{tabsecond}18.25&\cellcolor{tabsecond}0.690&0.226& 	\cellcolor{tabsecond}19.70&\cellcolor{tabsecond}0.738&\cellcolor{tabthird}0.189& 	\cellcolor{tabsecond}20.11&\cellcolor{tabsecond}0.757&\cellcolor{tabsecond}0.175 \\
Full & \cellcolor{tabfirst}18.52 & \cellcolor{tabfirst}0.705& \cellcolor{tabfirst}0.212& 	\cellcolor{tabfirst}19.90&\cellcolor{tabfirst}0.748&\cellcolor{tabfirst}0.180& 	\cellcolor{tabfirst}20.15&\cellcolor{tabfirst}0.766&\cellcolor{tabfirst}0.169 \\
\bottomrule
\end{tabular}
}
\end{table}

\noindent\textbf{Quantitative Results.}
Tabs.~\ref{tab:comp_nvs_megascenes} and~\ref{tab:comp_nvs_phototourism} report the PSNR, SSIM, and LPIPS of all methods on MegaScenes and Phototourism, respectively.
Since both our method and optimization-based methods require a reference image to specify the reconstruction appearance, we split each ground-truth image in half.
The left half is used as the reference input, while the right half is used to compute the metrics for the rendered results.
The results clearly demonstrate the superiority of our method in all metrics.
Moreover, as the number of input images increases, the performance of our method steadily improves, without the drastic fluctuations observed in WorldMirror.
Optimization-based methods (FSGS, GS-W, WildGaussian) produce the worst results due to a lack of sufficient scene images for optimization.
Feedforward methods, including AnySplat and WorldMirror, are hampered by the diversity of lighting conditions in the input images, resulting in appearances that are inconsistent with the target scene.

\begin{figure}[t!]
    \centering
    \includegraphics[width=1.0\linewidth]{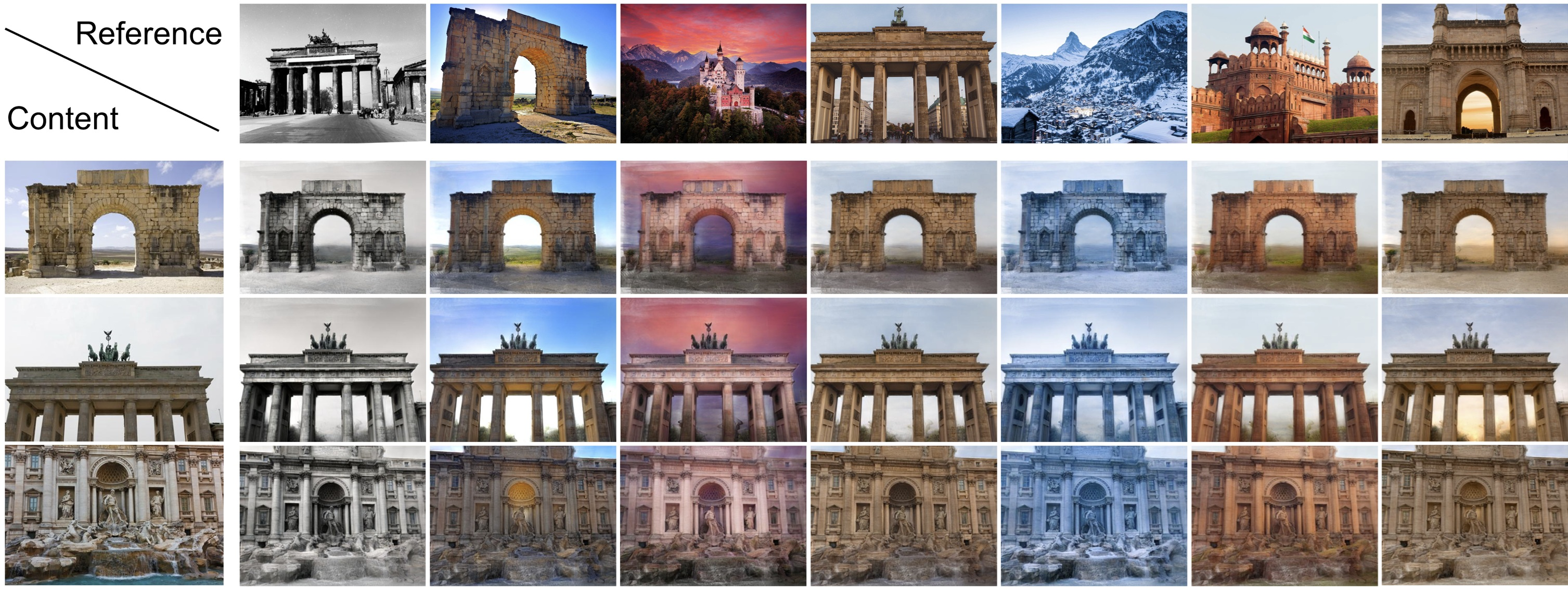}
    \caption{\textbf{Appearance transfer across diverse conditions}. The left column displays the target content scenes, and the top row provides various reference images captured under different lighting, weather, and times of day. Even when the references depict different building types, our model effectively transfers the overall color tone and appearance conditions while strictly preserving the underlying geometry.}
    \label{fig:appedit}
\end{figure}

    \setlength{\intextsep}{2pt} 
\setlength{\columnsep}{6pt} 

  \begin{wraptable}{r}{0.48\columnwidth}
    \centering
    \scriptsize
    \caption{\textbf{Pose evaluation.} We report the metrics on MegaScenes and highlight the best 
    and second-best results in \textbf{bold} and \underline{underline}, respectively.}
    \label{tab:pose}
    \setlength{\tabcolsep}{2.2pt}
    \renewcommand{\arraystretch}{0.95}
    \resizebox{\linewidth}{!}{%
    \begin{tabular}{lcccc}
      \toprule
      Method & AUC@5 $\uparrow$ & AUC@10 $\uparrow$ & AUC@30 $\uparrow$ & ATE $\downarrow$ \\
      \midrule
      VGGT    & \textbf{0.314} & \textbf{0.451} & \textbf{0.632} & \textbf{1.525} \\
      AnySplat & 0.249 & 0.392 & 0.596 & 1.564 \\
      Ours   & \underline{0.310} & \underline{0.443} & \underline{0.624} & \underline{1.559} \\
      \bottomrule
    \end{tabular}
    }
  \end{wraptable}
 
\noindent\textbf{Pose Evaluation.} 
We further evaluate camera pose estimation on 20 scenes sampled from MegaScenes, 
with each scene evaluated using 24 views.
As reported in~\cref{tab:pose}, we compare methods using AUC@5, AUC@10, AUC@30, 
and ATE metrics.
Our method improves over AnySplat across the AUC metrics and achieves results 
close to VGGT, a dedicated and strong pose prior.
These results show that our geometry branch is able to learn meaningful camera poses.

\subsection{Ablation Study}
To evaluate the design choices of WildSplat, we conduct an ablation study on the Phototourism dataset~\cite{snavely2006photo} using 4-, 8-, and 12-view settings, with quantitative results summarized in~\cref{tab:ablation}.
Decoupling geometry from appearance is essential for in-the-wild scenes. 
Removing the joint multi-reference training strategy (w/o Multi-Ref Supervision) results in the most significant performance drop across all metrics. Using only a single reference view per iteration tends to entangle the 3D geometry with specific lighting conditions, causing unstable optimization during training. By simultaneously rendering multiple appearance conditions, our method maintains training stability and disentangles the underlying geometry and appearance.
Furthermore, traditional pose-ranking methods often struggle with the extreme variations in viewpoint and focal length typical of unconstrained photo collections. By explicitly accounting for spatial overlap and scale consistency, the geometry-guided sampling provides the network with more informative and reliable context views, significantly enhancing the quality of novel view synthesis.
Finally, omitting the global modulation scheme (w/o Pre-Modulation) degrades performance, demonstrating that using the global \texttt{[CLS]} token to pre-modulate intermediate content features provides an appearance prior that reduces interference from inconsistent appearance in the content features. 
Our full model achieves the best performance, indicating that these components work together to enable robust, appearance-conditioned synthesis from sparse inputs.

\begin{figure}[t]
    \centering
    \includegraphics[width=1.0\linewidth]{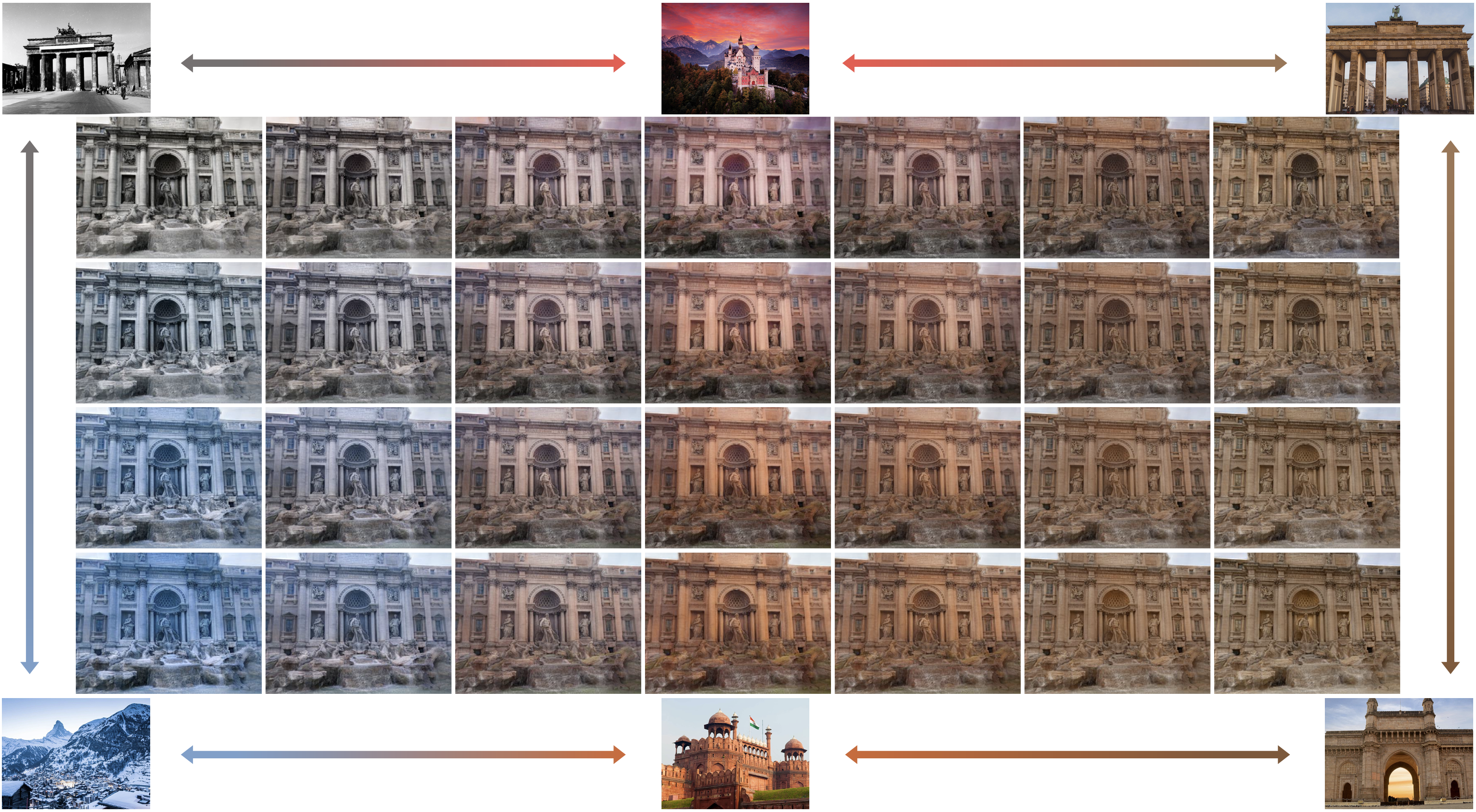}
    \caption{\textbf{Appearance interpolation}. The grid demonstrates continuous interpolation across distinct reference appearances. By blending the extracted features, we achieve faithful and seamless transitions between the source images.}
    \label{fig:interp}
\end{figure}
\subsection{Appearance Editing}

Since our method factorizes the reconstruction process into illumination-agnostic geometry extraction and appearance-aware conditioning, we can adjust the reference image to inject different appearance features into the same scene. 
As shown in Fig.~\ref{fig:appedit}, we present qualitative results of our method using reference images under varying weather conditions, times of day, and even across different building types.
Our method successfully captures the overall color tone from the reference image and transfers it to the target scene. 
Thanks to the decoupled geometry and color modeling architecture, combined with a multi-reference training paradigm tailored for photometrically inconsistent data, our method can distinguish between the background and the main building in the reference image, accurately transferring their colors to the corresponding regions in the target scene.
\cref{fig:interp} further illustrates appearance interpolation between given reference images. By interpolating the extracted features, we achieve smooth visual blending, resulting in seamless transitions across different lighting and style conditions. Finally, our feedforward model eliminates the need for redundant per-scene optimization, making the approach highly efficient and practical for real-world scene style transfer tasks.
\section{Conclusion}
\label{sec:conclusion}

In this work, we present WildSplat, the first feedforward 3D Gaussian Splatting framework 
designed for unposed in-the-wild images exhibiting severe appearance variations. 
To address the inherent photometric inconsistencies of such data, 
we propose a dual-branch architecture that explicitly decouples geometry from appearance. 
The geometry branch leverages a VGGT-based backbone to jointly estimate camera poses 
and extract an illumination-agnostic, shared 3D geometry. 
Our appearance branch employs an appearance injector module that utilizes 
feature pre-modulation and cross-attention to seamlessly inject target 
reference information into the content features. 
Extensive experiments demonstrate that WildSplat achieves state-of-the-art 
photorealistic novel view synthesis for scenes with various appearances, 
while also enabling faithful appearance editing. 
 

\section*{Acknowledgements}
This work was partially supported by NSF of China (No. 62425209).
%
%
\bibliographystyle{splncs04}
\bibliography{main}

@String(PAMI  = {IEEE Trans. Pattern Anal. Mach. Intell.})

@String(IJCV  = {Int. J. Comput. Vis.})

@String(CVPR  = {IEEE Conf. Comput. Vis. Pattern Recog.})

@String(ICCV  = {Int. Conf. Comput. Vis.})

@String(ECCV  = {Eur. Conf. Comput. Vis.})

@String(NeurIPS = {Adv. Neural Inform. Process. Syst.})

@String(TOG   = {ACM Trans. Graph.})

@inproceedings{martin2021nerf,
  title={{NeRF} in the Wild: Neural Radiance Fields for Unconstrained Photo Collections},
  author={Martin-Brualla, Ricardo and Radwan, Noha and Sajjadi, Mehdi SM and Barron, Jonathan T and Dosovitskiy, Alexey and Duckworth, Daniel},
  booktitle=CVPR,
  pages={7210--7219},
  year={2021}
}

@inproceedings{zhang2024gaussian,
  title={Gaussian in the Wild: {3D} Gaussian Splatting for Unconstrained Image Collections},
  author={Zhang, Dongbin and Wang, Chuming and Wang, Weitao and Li, Peihao and Qin, Minghan and Wang, Haoqian},
  booktitle=ECCV,
  pages={341--359},
  year={2024},
  organization={Springer}
}

@inproceedings{kulhanek2024wildgaussians,
  title={{W}ild{G}aussians: {3D} Gaussian Splatting in the Wild},
  author={Kulhanek, Jonas and Peng, Songyou and Kukelova, Zuzana and Pollefeys, Marc and Sattler, Torsten},
  booktitle=NeurIPS,
  year={2024}
}

@article{oquab2023dinov2,
  title={{DINOv2}: Learning Robust Visual Features without Supervision},
  author={Oquab, Maxime and Darcet, Timoth{\'e}e and Moutakanni, Th{\'e}o and Vo, Huy and Szafraniec, Marc and Khalidov, Vasil and Fernandez, Pierre and Haziza, Daniel and Massa, Francisco and El-Nouby, Alaaeldin and others},
  journal={arXiv preprint arXiv:2304.07193},
  year={2023}
}

@article{xu2024wild,
  title={{Wild-GS}: Real-Time Novel View Synthesis from Unconstrained Photo Collections},
  author={Xu, Jiacong and Mei, Yiqun and Patel, Vishal},
  journal=NeurIPS,
  volume={37},
  pages={103334--103355},
  year={2024}
}

@article{xu2024splatfacto,
  title={{Splatfacto-W}: A {Nerfstudio} Implementation of Gaussian Splatting for Unconstrained Photo Collections},
  author={Xu, Congrong and Kerr, Justin and Kanazawa, Angjoo},
  journal={arXiv preprint arXiv:2407.12306},
  year={2024}
}

@inproceedings{schonberger2016structure,
  title={Structure-from-motion revisited},
  author={Schonberger, Johannes L and Frahm, Jan-Michael},
  booktitle=CVPR,
  pages={4104--4113},
  year={2016}
}

@inproceedings{yu2021pixelnerf,
  title={{pixelNeRF}: Neural Radiance Fields from One or Few Images},
  author={Yu, Alex and Ye, Vickie and Tancik, Matthew and Kanazawa, Angjoo},
  booktitle=CVPR,
  pages={4578--4587},
  year={2021}
}

@inproceedings{chen2021mvsnerf,
  title={{MVSNeRF}: Fast Generalizable Radiance Field Reconstruction from Multi-View Stereo},
  author={Chen, Anpei and Xu, Zexiang and Zhao, Fuqiang and Zhang, Xiaoshuai and Xiang, Fanbo and Yu, Jingyi and Su, Hao},
  booktitle=ICCV,
  pages={14124--14133},
  year={2021}
}

@article{chen2025explicit,
  title={Explicit correspondence matching for generalizable neural radiance fields},
  author={Chen, Yuedong and Xu, Haofei and Wu, Qianyi and Zheng, Chuanxia and Cham, Tat-Jen and Cai, Jianfei},
  journal=PAMI,
  year={2025},
  doi={10.1109/TPAMI.2025.3598711}
}

@inproceedings{du2023learning,
  title={Learning to render novel views from wide-baseline stereo pairs},
  author={Du, Yilun and Smith, Cameron and Tewari, Ayush and Sitzmann, Vincent},
  booktitle=CVPR,
  pages={4970--4980},
  year={2023}
}

@article{miyato2023gta,
  title={{GTA}: A Geometry-Aware Attention Mechanism for Multi-View Transformers},
  author={Miyato, Takeru and Jaeger, Bernhard and Welling, Max and Geiger, Andreas},
  journal={arXiv preprint arXiv:2310.10375},
  year={2023}
}

@inproceedings{sajjadi2022scene,
  title={Scene representation transformer: Geometry-free novel view synthesis through set-latent scene representations},
  author={Sajjadi, Mehdi SM and Meyer, Henning and Pot, Etienne and Bergmann, Urs and Greff, Klaus and Radwan, Noha and Vora, Suhani and Lu{\v{c}}i{\'c}, Mario and Duckworth, Daniel and Dosovitskiy, Alexey and others},
  booktitle=CVPR,
  pages={6229--6238},
  year={2022}
}

@inproceedings{xu2024murf,
  title={{MuRF}: Multi-Baseline Radiance Fields},
  author={Xu, Haofei and Chen, Anpei and Chen, Yuedong and Sakaridis, Christos and Zhang, Yulun and Pollefeys, Marc and Geiger, Andreas and Yu, Fisher},
  booktitle=CVPR,
  pages={20041--20050},
  year={2024}
}

@inproceedings{charatan2024pixelsplat,
  title={{pixelSplat}: {3D} Gaussian Splats from Image Pairs for Scalable Generalizable {3D} Reconstruction},
  author={Charatan, David and Li, Sizhe Lester and Tagliasacchi, Andrea and Sitzmann, Vincent},
  booktitle=CVPR,
  pages={19457--19467},
  year={2024}
}

@inproceedings{chen2024mvsplat,
  title={{MVSplat}: Efficient {3D} Gaussian Splatting from Sparse Multi-View Images},
  author={Chen, Yuedong and Xu, Haofei and Zheng, Chuanxia and Zhuang, Bohan and Pollefeys, Marc and Geiger, Andreas and Cham, Tat-Jen and Cai, Jianfei},
  booktitle=ECCV,
  pages={370--386},
  year={2024},
  organization={Springer}
}

@article{ye2024noposplat,
      title   = {{NoPoSplat}: No Pose, No Problem: Surprisingly Simple {3D} Gaussian Splats from Sparse Unposed Images},
      author  = {Ye, Botao and Liu, Sifei and Xu, Haofei and Xueting, Li and Pollefeys, Marc and Yang, Ming-Hsuan and Songyou, Peng},
      journal = {arXiv preprint arXiv:2410.24207},
      year    = {2024}
    }

@article{jiang2025anysplat,
  title={{AnySplat}: Feed-Forward {3D} Gaussian Splatting from Unconstrained Views},
  author={Jiang, Lihan and Mao, Yucheng and Xu, Linning and Lu, Tao and Ren, Kerui and Jin, Yichen and Xu, Xudong and Yu, Mulin and Pang, Jiangmiao and Zhao, Feng and others},
  journal=TOG,
  volume={44},
  number={6},
  pages={1--16},
  year={2025},
  publisher={ACM New York, NY, USA}
}

@inproceedings{wang2025vggt,
  title={{VGGT}: Visual Geometry Grounded Transformer},
  author={Wang, Jianyuan and Chen, Minghao and Karaev, Nikita and Vedaldi, Andrea and Rupprecht, Christian and Novotny, David},
  booktitle=CVPR,
  pages={5294--5306},
  year={2025}
}

@inproceedings{peebles2023scalable,
  title={Scalable diffusion models with transformers},
  author={Peebles, William and Xie, Saining},
  booktitle=ICCV,
  pages={4195--4205},
  year={2023}
}

@inproceedings{dahmani2024swag,
  title={{SWAG}: Splatting in the Wild Images with Appearance-Conditioned Gaussians},
  author={Dahmani, Hiba and Bennehar, Moussab and Piasco, Nathan and Roldao, Luis and Tsishkou, Dzmitry},
  booktitle=ECCV,
  pages={325--340},
  year={2024},
  organization={Springer}
}

@article{liu2025worldmirror,
  title={{WorldMirror}: Universal {3D} World Reconstruction with Any-Prior Prompting},
  author={Liu, Yifan and Min, Zhiyuan and Wang, Zhenwei and Wu, Junta and Wang, Tengfei and Yuan, Yixuan and Luo, Yawei and Guo, Chunchao},
  journal={arXiv preprint arXiv:2510.10726},
  year={2025}
}

@inproceedings{waechter2014let,
  title={Let there be color! Large-scale texturing of 3D reconstructions},
  author={Waechter, Michael and Moehrle, Nils and Goesele, Michael},
  booktitle=ECCV,
  pages={836--850},
  year={2014},
  organization={Springer}
}

@inproceedings{1996photographs,
  author       = {Paul E. Debevec and
                  Camillo J. Taylor and
                  Jitendra Malik},
  title        = {Modeling and Rendering Architecture from Photographs: {A} Hybrid Geometry-
                  and Image-Based Approach},
  booktitle    = {SIGGRAPH},
  pages        = {11--20},
  publisher    = {{ACM}},
  year         = {1996},
}

@inproceedings{liu2019soft,
  title={Soft rasterizer: A differentiable renderer for image-based 3d reasoning},
  author={Liu, Shichen and Li, Tianye and Chen, Weikai and Li, Hao},
  booktitle=ICCV,
  pages={7708--7717},
  year={2019}
}

@article{kutulakos2000theory,
  title={A theory of shape by space carving},
  author={Kutulakos, Kiriakos N and Seitz, Steven M},
  journal=IJCV,
  volume={38},
  number={3},
  pages={199--218},
  year={2000},
  publisher={Springer}
}

@inproceedings{wang2024dust3r,
  title={{DUSt3R}: Geometric {3D} Vision Made Easy},
  author={Wang, Shuzhe and Leroy, Vincent and Cabon, Yohann and Chidlovskii, Boris and Revaud, Jerome},
  booktitle=CVPR,
  pages={20697--20709},
  year={2024}
}

@article{seitz1999photorealistic,
  title={Photorealistic scene reconstruction by voxel coloring},
  author={Seitz, Steven M and Dyer, Charles R},
  journal=IJCV,
  volume={35},
  number={2},
  pages={151--173},
  year={1999},
  publisher={Springer}
}

@article{szeliski1999stereo,
  title={Stereo matching with transparency and matting},
  author={Szeliski, Richard and Golland, Polina},
  journal=IJCV,
  volume={32},
  number={1},
  pages={45--61},
  year={1999},
  publisher={Springer}
}

@article{mildenhall2021nerf,
  title={Nerf: Representing scenes as neural radiance fields for view synthesis},
  author={Mildenhall, Ben and Srinivasan, Pratul P and Tancik, Matthew and Barron, Jonathan T and Ramamoorthi, Ravi and Ng, Ren},
  journal={Communications of the ACM},
  volume={65},
  number={1},
  pages={99--106},
  year={2021},
  publisher={ACM New York, NY, USA}
}

@article{kerbl20233d,
  title={{3D} Gaussian Splatting for Real-Time Radiance Field Rendering},
  author={Kerbl, Bernhard and Kopanas, Georgios and Leimk{\"u}hler, Thomas and Drettakis, George and others},
  journal=TOG,
  volume={42},
  number={4},
  pages={139--1},
  year={2023}
}

@article{zhai2025splatloc,
  title={{SplatLoc}: {3D} Gaussian Splatting-Based Visual Localization for Augmented Reality},
  author={Zhai, Hongjia and Zhang, Xiyu and Zhao, Boming and Li, Hai and He, Yijia and Cui, Zhaopeng and Bao, Hujun and Zhang, Guofeng},
  journal={arXiv preprint arXiv:2409.14067},
  year={2024}
}

@article{zhang2026atlasgs,
  title={{AtlasGS}: Atlanta-World Guided Surface Reconstruction with Implicit Structured Gaussians},
  author={Zhang, Xiyu and Bao, Chong and Chen, Yipeng and Zhai, Hongjia and Dong, Yitong and Bao, Hujun and Cui, Zhaopeng and Zhang, Guofeng},
  journal={arXiv preprint arXiv:2510.25129},
  year={2025},
  note={{NeurIPS} 2025}
}

@article{zhai2026onlinepg,
  title={{OnlinePG}: Online Open-Vocabulary Panoptic Mapping with {3D} Gaussian Splatting},
  author={Zhai, Hongjia and Zhang, Qi and Pan, Xiaokun and Zhang, Xiyu and Dong, Yitong and Zhang, Huaqi and Xu, Dan and Zhang, Guofeng},
  journal={arXiv preprint arXiv:2603.18510},
  year={2026},
  note={{CVPR} 2026}
}

@inproceedings{barron2021mip,
  title={{Mip-NeRF}: A Multiscale Representation for Anti-Aliasing Neural Radiance Fields},
  author={Barron, Jonathan T and Mildenhall, Ben and Tancik, Matthew and Hedman, Peter and Martin-Brualla, Ricardo and Srinivasan, Pratul P},
  booktitle=ICCV,
  pages={5855--5864},
  year={2021}
}

@inproceedings{barron2022mip,
  title={{Mip-NeRF 360}: Unbounded Anti-Aliased Neural Radiance Fields},
  author={Barron, Jonathan T and Mildenhall, Ben and Verbin, Dor and Srinivasan, Pratul P and Hedman, Peter},
  booktitle=CVPR,
  pages={5470--5479},
  year={2022}
}

@inproceedings{barron2023zip,
  title={{Zip-NeRF}: Anti-Aliased Grid-Based Neural Radiance Fields},
  author={Barron, Jonathan T and Mildenhall, Ben and Verbin, Dor and Srinivasan, Pratul P and Hedman, Peter},
  booktitle=ICCV,
  pages={19697--19705},
  year={2023}
}

@inproceedings{ling2024dl3dv,
  title={{DL3DV-10K}: A Large-Scale Scene Dataset for Deep Learning-Based {3D} Vision},
  author={Ling, Lu and Sheng, Yichen and Tu, Zhi and Zhao, Wentian and Xin, Cheng and Wan, Kun and Yu, Lantao and Guo, Qianyu and Yu, Zixun and Lu, Yawen and others},
  booktitle=CVPR,
  pages={22160--22169},
  year={2024}
}

@inproceedings{tung2024megascenes,
  title={{MegaScenes}: Scene-Level View Synthesis at Scale},
  author={Tung, Joseph and Chou, Gene and Cai, Ruojin and Yang, Guandao and Zhang, Kai and Wetzstein, Gordon and Hariharan, Bharath and Snavely, Noah},
  booktitle=ECCV,
  pages={197--214},
  year={2024},
  organization={Springer}
}

@inproceedings{li2018megadepth,
  title={{MegaDepth}: Learning Single-View Depth Prediction from Internet Photos},
  author={Li, Zhengqi and Snavely, Noah},
  booktitle=CVPR,
  pages={2041--2050},
  year={2018}
}

@inproceedings{zhu2024fsgs,
  title={{FSGS}: Real-Time Few-Shot View Synthesis Using Gaussian Splatting},
  author={Zhu, Zehao and Fan, Zhiwen and Jiang, Yifan and Wang, Zhangyang},
  booktitle=ECCV,
  pages={145--163},
  year={2024},
  organization={Springer}
}

@article{snavely2006photo,
  author       = {Noah Snavely and
                  Steven M. Seitz and
                  Richard Szeliski},
  title        = {Photo tourism: exploring photo collections in 3D},
  journal      = TOG,
  volume       = {25},
  number       = {3},
  pages        = {835--846},
  year         = {2006},
}

@article{liu2025tc,
  title={{TC-Light}: Temporally Coherent Generative Rendering for Realistic World Transfer},
  author={Liu, Yang and Luo, Chuanchen and Tang, Zimo and Li, Yingyan and Yang, Yuran and Ning, Yuanyong and Fan, Lue and Peng, Junran and Zhang, Zhaoxiang},
  journal={arXiv preprint arXiv:2506.18904},
  year={2025}
}

\clearpage
\title{Supplementary Material for\\
WildSplat: Feedforward Gaussian Splatting from Unposed In-the-Wild Images}
\titlerunning{Supplementary Material}
\author{}
\institute{}
\maketitle
\appendix
\setcounter{figure}{0}
\setcounter{table}{0}
\renewcommand{\thefigure}{\Alph{figure}}
\renewcommand{\thetable}{\Alph{table}}
\renewcommand{\theHfigure}{supp.figure.\arabic{figure}}
\renewcommand{\theHtable}{supp.table.\arabic{table}}

\afterpage{
\begin{center}
    \centering
    \includegraphics[width=0.88\linewidth]{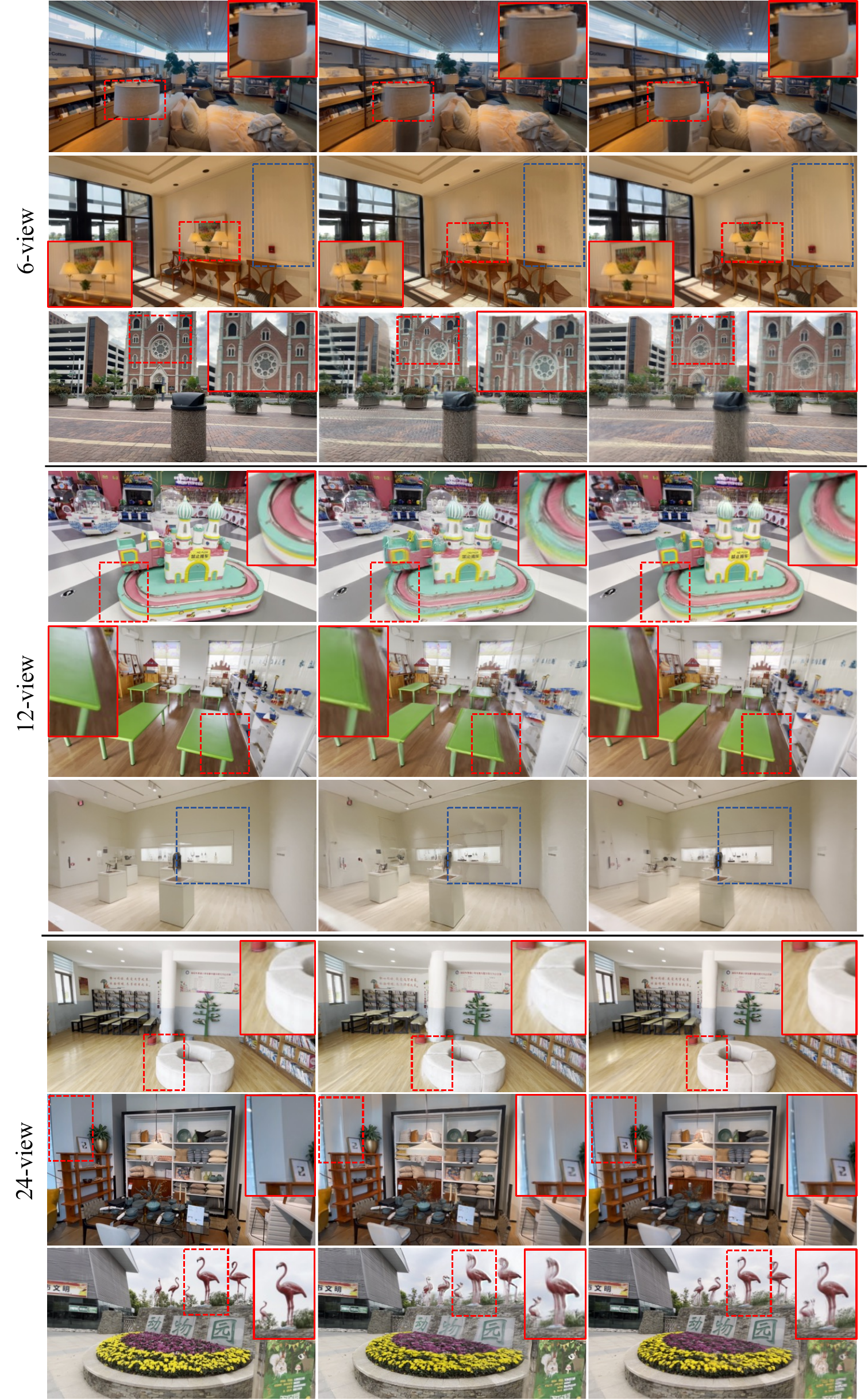}
    \makebox[0.88\linewidth][r]{%
        \makebox[0.273\linewidth][c]{\small GT}%
        \makebox[0.273\linewidth][c]{\small AnySplat\rlap{~\cite{jiang2025anysplat}}}%
        \makebox[0.273\linewidth][c]{\small \textbf{Ours}}%
    }
    \captionof{figure}{\textbf{Qualitative results} on DL3DV under different sparse-view settings. Red boxes highlight zoomed-in details for visual comparison of reconstruction quality, while blue dashed boxes indicate artifacts caused by view-inconsistent appearances.}
    \label{fig:dl3dv}
\end{center}
\newpage
\begin{center}
    \centering
    \small
    \setlength{\tabcolsep}{4pt}
    \renewcommand{\arraystretch}{1.1}
    \captionof{table}{\textbf{Quantitative comparison} of novel view synthesis on the DL3DV benchmark. Performance is measured using PSNR ($\uparrow$), SSIM ($\uparrow$), and LPIPS ($\downarrow$). Our method demonstrates superior reconstruction quality and consistency across all settings.}
    \label{tab:comp_nvs_dl3dv}
    \resizebox{0.9\linewidth}{!}{
    \begin{tabular}{@{}lccccccccc@{}}
    \toprule
    Method & \multicolumn{3}{c}{6 views} & \multicolumn{3}{c}{12 views} & \multicolumn{3}{c}{24 views} \\
    \midrule
     & PSNR$\uparrow$ & SSIM$\uparrow$ & LPIPS$\downarrow$ & PSNR$\uparrow$ & SSIM$\uparrow$ & LPIPS$\downarrow$ & PSNR$\uparrow$ & SSIM$\uparrow$ & LPIPS$\downarrow$ \\
    \midrule
    AnySplat~\cite{jiang2025anysplat} & 20.33&0.643&0.154 &	19.73&0.603&0.187 &19.55 &0.589&	0.202\\
    Ours & \textbf{22.37}&\textbf{0.752}&\textbf{0.131} &	\textbf{21.10}&\textbf{0.686}&\textbf{0.167} & \textbf{20.67} & \textbf{0.666} &\textbf{0.185} \\
    \bottomrule
    \end{tabular}
    }
\end{center}
}

\section{Extended Implementation Details}
\label{sec:detail}
\subsection{Network Architecture}

The geometry branch is instantiated with a DINOv2-large encoder~\cite{oquab2023dinov2} ($D=1024$) followed by $L=24$ alternating-attention layers~\cite{wang2025vggt}. For appearance extraction, we adopt a pretrained DINOv2-base model~\cite{oquab2023dinov2}. The Appearance Injector comprises four identical blocks, each containing an AdaLN-Zero modulation layer, a cross-attention layer, a self-attention layer, and an MLP. The network predicts pixel-aligned 3D Gaussians, parameterizing view-dependent colors with spherical harmonics of degree $k=4$. All parameters are optimized except for the geometry encoder. To optimize memory consumption during training and inference, we employ a voxel-based merging strategy~\cite{jiang2025anysplat} to aggregate spatially redundant Gaussians.

\subsection{Training Details}

All models are optimized using the Adam optimizer ($\beta_1=0.9$, $\beta_2=0.95$) across four NVIDIA RTX H20 GPUs. The initial learning rate is set to $2 \times 10^{-5}$ for the network backbones, while a higher learning rate of $2 \times 10^{-4}$ is applied to the DPT heads. The loss balancing weights are empirically set to $\lambda_1=10.0$ for pose distillation, $\lambda_2=10.0$ for MSE, $\lambda_3=1.0$ for SSIM, and $\lambda_4=0.05$ for LPIPS.

During the training stage, the input images of a scene are partitioned into context and target views. The network jointly predicts camera poses for all views, but constructs the 3D Gaussians and content features strictly from the context views. 
To construct a batch, we dynamically sample from 2 to 16 context views for the DL3DV dataset~\cite{ling2024dl3dv}, and from 2 to 12 views for MegaScenes~\cite{tung2024megascenes}. All sampled images are resized and randomly cropped to a resolution of 448$\times$448.

To prevent the geometry branch from overfitting to specific lighting conditions, we apply color jittering to the input images, with brightness and saturation jittered by 0.3, contrast by 0.5, and a maximum hue shift of 0.1. 
Meanwhile, the reference images are selected from the input set and randomly cropped into left or right halves. 
This deliberate spatial mismatch encourages the appearance branch to capture global color statistics rather than spatially aligned layouts. Finally, for Geometry-guided View Sampling, we utilize sparse SfM points to enforce valid multi-view constraints. The thresholds are empirically set as follows: spatial overlap $\tau_s=0.25$, scale consistency $\tau_r=1.3$, and minimum point count $\tau_n=5$, where $N$ is the number of context views.

\subsection{Synthetic Multi-illumination Data.} 
To construct paired multi-illumination data for Stage 2 training, we augment a subset of the DL3DV dataset. Specifically, we randomly sample 600 video sequences and employ an LLM to formulate a detailed scene description for each. Based on this description, the LLM is further prompted to generate diverse lighting-editing text prompts. Finally, an off-the-shelf video editing model~\cite{liu2025tc} is used to relight the original videos based on the generated prompts. Through this automated pipeline, we synthesize 5 distinct lighting conditions per video sequence.

\section{Extended Experiments}
\label{sec:add_exps}

\subsection{Novel View Synthesis}
Beyond in-the-wild synthesis, our decoupled architecture remains effective for scenes with consistent illumination. We validate this on the DL3DV benchmark~\cite{ling2024dl3dv}, which features 140 photometrically consistent indoor and outdoor scenes.

As shown in~\cref{tab:comp_nvs_dl3dv}, WildSplat consistently outperforms the baseline AnySplat~\cite{jiang2025anysplat} across sparse-view configurations of 6, 12, and 24 views with maximum frame intervals of 50, 100, and 150. In these experiments, we maintain a maximum image length of 448, resulting in an input and output resolution of 448$\times$252.
Specifically, our method achieves higher PSNR and SSIM scores while maintaining a lower LPIPS error across all tested settings. 
The qualitative comparisons in~\cref{fig:dl3dv} further illustrate the advantages of our method. 
AnySplat struggles with view-inconsistent appearances in the input images, leading to noticeable rendering artifacts indicated by blue dashed boxes. 
By explicitly decoupling geometry and appearance, WildSplat successfully suppresses these artifacts, reconstructing better geometric details highlighted by red boxes and maintaining strict consistency with the designated reference appearance.

\subsection{Computational Efficiency}

\begin{figure}[t]
    \centering
    \includegraphics[width=1.0\linewidth]{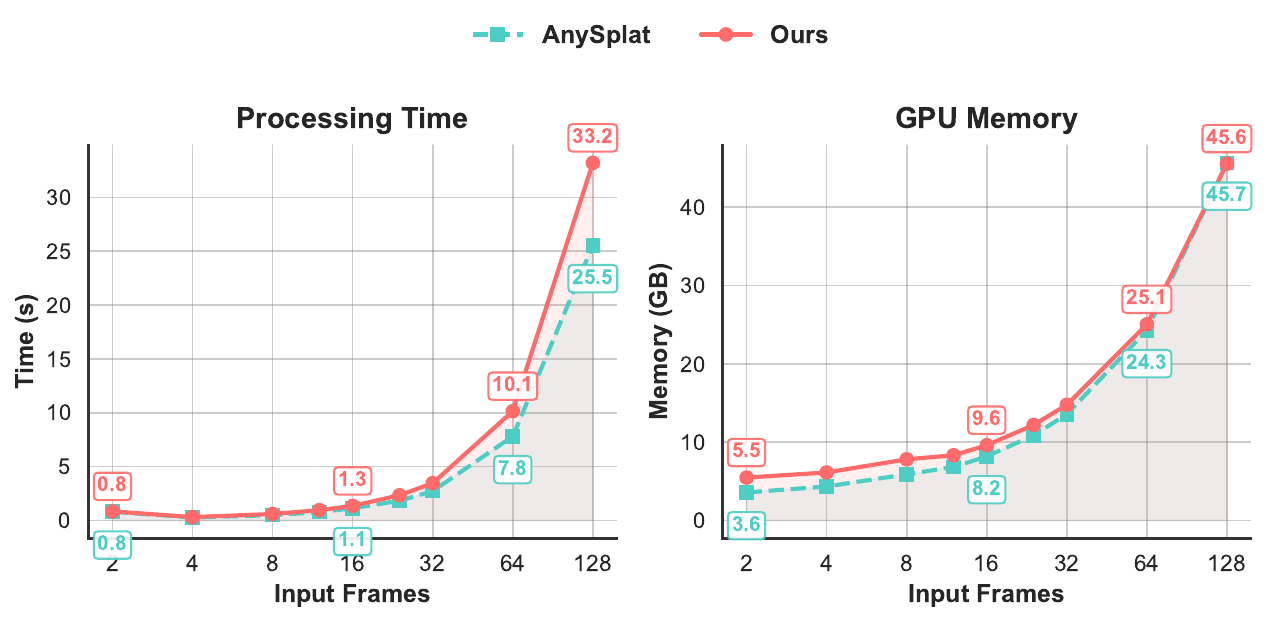}
    \caption{
        \textbf{Computational efficiency comparison.} We report the processing time (left) and GPU memory consumption (right) under varying input views.
    }
    \label{fig:performance}
\end{figure}

We compare the computational efficiency of our method against AnySplat by measuring the processing time and peak GPU memory consumption during a single feedforward pass. 
The evaluation is conducted across varying numbers of input frames, ranging from 2 to 128. For this performance comparison, we utilize a fixed resolution of 448$\times$448 for all inputs. 
As shown in~\cref{fig:performance}, WildSplat achieves competitive efficiency relative to AnySplat.
For a standard 16-view input, our method requires only 1.3 seconds of processing time and 9.6 GB of GPU memory, adding minimal overhead compared to AnySplat's 1.1 seconds and 8.2 GB. 
When scaling up to an extreme 128 input frames, the memory footprint remains virtually identical (45.6 GB for Ours vs. 45.7 GB for AnySplat), with a modest, acceptable increase in processing time (33.2 seconds vs. 25.5 seconds). 
This demonstrates that our Appearance Branch and global pre-modulation operations introduce negligible computational cost while enabling robust appearance-conditioned synthesis.

\begin{figure}[t]
    \centering
    \includegraphics[width=0.9\linewidth]{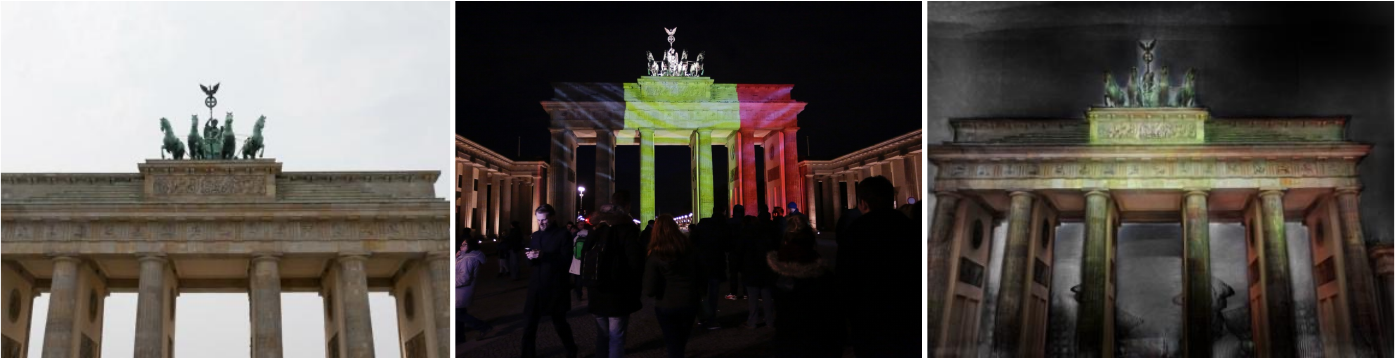}
    \begin{tabular}{@{}c@{}c@{}c@{}}
        \makebox[0.32\linewidth]{\small Content} &
        \makebox[0.32\linewidth]{\small Reference} &
        \makebox[0.32\linewidth]{\small Ours}
    \end{tabular}
    \caption{\textbf{Failure case of WildSplat.} Localized projected illumination may not be faithfully preserved and can be blended into nearby surfaces.}
    \label{fig:failure_cases}
\end{figure}

\subsection{Failure Cases \& Limitations}
WildSplat remains limited in handling highly localized, spatially varying illumination.
As shown in~\cref{fig:failure_cases}, projected light patterns may be missed or incorrectly blended into nearby surfaces, since such effects are not consistently tied to the underlying scene geometry.
Another limitation is scalability.
The alternating-attention layers and dense 3D Gaussian prediction increase memory usage with image resolution and the number of context views, limiting deployment on consumer-grade GPUs for extreme-scale scenes.
Future work may explore spatially localized appearance modeling for complex lighting effects, as well as sparse attention or hierarchical refinement to reduce computational overhead.

\end{document}